# Non-destructive Testing of Composite Fibre Materials with Hyperspectral Imaging – Evaluative Studies in the EU H2020 FibreEUse Project

Yijun Yan[1], *Member, IEEE*, Jinchang Ren[1*], *Senior Member, IEEE*, Huan Zhao[2], James F.C. Windmill[2], *Senior Member, IEEE*, Winifred Ijomah[3], Jesper de Wit[4], and Justus von Freeden[5]

[1] National Subsea Centre, Robert Gordon University, Aberdeen, U.K.
[2] Dept. of Electronic and Electrical Engineering, University of Strathclyde, Glasgow, U.K.
[3] Dept of Design, Manufacturing & Engineering Management, University of Strathclyde, Glasgow, U.K.
[4] INVENT GmbH, Christian-Pommer-Straße 47, 38112 Braunschweig, Germany
[5] Fraunhofer Institute for Machine Tools and Forming Technology IWU, Hermann-Münch-Str. 1, 38440 Wolfsburg, Germany

*Abstract*: Through capturing spectral data from a wide frequency range along with the spatial information, hyperspectral imaging (HSI) can detect minor differences in terms of temperature, moisture and chemical composition. Therefore, HSI has been successfully applied in various applications, including remote sensing for security and defense, precision agriculture for vegetation and crop monitoring, food/drink, and pharmaceuticals quality control. However, for condition monitoring and damage detection in carbon fibre reinforced polymer (CFRP), the use of HSI is a relatively untouched area, as existing non-destructive testing (NDT) techniques focus mainly on delivering information about physical integrity of structures but not on material composition. To this end, HSI can provide a unique way to tackle this challenge. In this paper, with the use of a near-infrared HSI camera, applications of HSI for the non-destructive inspection of CFRP products are introduced, taking the EU H2020 FibreEUse project as the background. Technical challenges and solutions on three case studies are presented in detail, including adhesive residues detection, surface damage detection and Cobot based automated inspection. Experimental results have fully demonstrated the great potential of HSI and related vision techniques for NDT of CFRP, especially the potential to satisfy the industrial manufacturing environment.

## 1 INTRODUCTION

Due to the unique characteristics of lightweight, high stiffness/strength and damping resistant, carbon fibre reinforced polymer composites (CFRP) has been widely used as structural materials in many areas such as aerospace, marine, transport, sports and civil engineering industries [1, 2], as illustrated in Fig. 1(a). Composites are relatively young and intrinsically durable materials, but composite-based components or products have a limited lifetime, which is usually less than 20-30 years, e.g. 20-25 years for a wind turbine [3], and ~10 years on average for recreational boats and car bodies [4]. Considering the continuously and quickly rising demand for composites in industrial manufacturing, correct waste management of End-of-life (EoL) CFRP becomes urgent and vital. Landfilling of EoL composites is still the most popular waste management strategy, which is a relatively cheap option, but it is not the most optimal way and doesn't comply with the Waste Framework Directive (2008/98/EC). In the future, the landfilling will become unviable because of the higher legislation-driven cost. Several countries, e.g., Germany and Austria, have already forbidden landfilling of composite waste, and other EU countries will follow suit soon.

For the EoL composites, remanufacturing and reuse is the best solution, which can not only comply with environmental regulations, but also benefit both end users and stake holders. However, many barriers have affected the application of this solution, such as the negative perception of recycled products, lack of suitable business models, immature recycling techniques, limited synergistic use of available inspection, and repair and reprocessing technologies et al. To address all these challenges, an EU H2020 project, FibreEUse, is funded, aiming to develop effective recycling, remanufacturing, inspection solutions and profitable reuse options for EoL

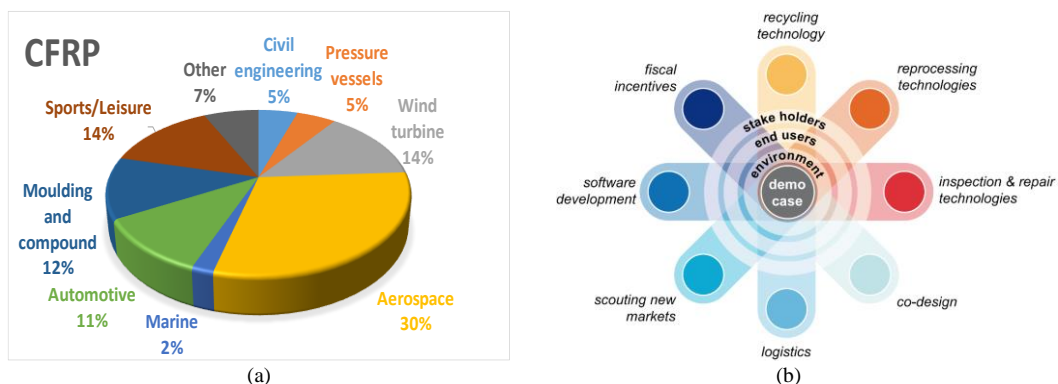

Fig. 1. (a) CFRP production in Europe for different industries, (b) the concept of FibreEUse, source: http://fibreeuse.eu/

composites, see in Fig. 1(b), through the integration of innovative remanufacturing technologies.

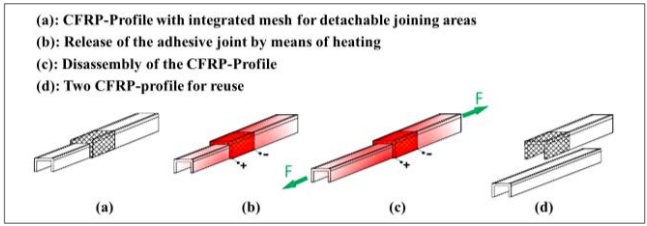

Fig. 2. Illustration of separation of bonded parts.

During the remanufacturing process, the EoL products will be disjointed into individual components (Fig. 2), which are then refurbished and reused in new products afterwards. As one of the most important modules in this workflow, non-destructive testing is crucially required to provide a reliable and efficient inspection to assess the condition of the composites in the remanufacturing chain.

Existing non-intrusive inspection techniques include visual inspection, ultrasonic testing, and thermal imaging [5]. Visual inspection is performed in optical wavelengths that are suitable for the human eyes, which is a fast and affordable inspection method as it doesn't require additional NDT instruments. However, it is labour intensive, and unreliable for small defects compared to an instrumented inspection. To tackle these difficulties, Red (R), Green (G) and Blue (B) based RGB colour camera was used for improved damage detection and fault localization [6]. However, its performance can still be easily affected by various weather and lighting conditions [7]. Ultrasonic testing can detect the internal flaws by analysing the sending and the receiving pulsed echo between one and/or multiple transducer(s). It is an efficient and safe technique, but it sometimes suffers in dealing with non-smooth and complex geometry surfaces, pretentious to subjective and objective factors [8]. Thermal imaging is a thermoelastic effect based inspection technique, where defects in the testing region is expected to present a higher temperature response [9]. This technique requires sensitive and expensive instrumentation, and highly skilled technician to run the instruments [5]. Moreover, the conduction and convection processes may also affect the temperature variations, leading to a barrier to localise the inspection [10]. In general, these techniques mostly extract information about physical integrity of the structures rather than the material composition from the objects. To this end, HSI has provided a unique solution to tackle this challenge.

HSI is actually an emerging technique that integrates imaging technique and spectroscopy, enabling the acquisition of spectral data from a wide frequency range along with the spatial information. Thus, it can detect minor differences in terms of temperature, moisture and chemical composition, making it a unique solution for nonintrusive inspection far beyond conventional techniques [11]. HSI has been applied successfully in wide range of applications such as food and drinks [12, 13], remote sensing[14, 15], clinical medicine [16], and art verification [17], etc. However, its application in (re)manufacturing is rare, except some limited works for plastic characterisation [18, 19] and defect detection from wind turbine blades [8]. As an emerging application, applying HSI for non-destructive inspection on fibre composites is presented in this paper, where three case studies are covered, including adhesives residues detection, surface damage detection, and robot integrated industrial inspection.

FRP products usually comprise FRP components or a combination of CFRP and metal components via adhesive bonding. For reuse purpose, these EoL FRP products need be disassembled with the adhesive bond being separated. However, adhesive residues will still adhere to the surface of the components and structures, which will be cleaned by laser treatment whilst the surface will be refurbished before the next bond. Although laser treatment has the potential advantage to clean the remnants adhesive, it has the risk of thermally degradation of the joint section. To monitor the surface condition such as the residual adhesive, and grinding defect caused during the laser treatment in real time, an NDT method is highly needed. To tackle with this issue, HSI will be introduced, and its usefulness will be demonstrated as our first case study.

After the process of adhesive removal, the condition of each EoL component is decisive for their reuse. Components with irreparable damages cannot be reused and thus need to be recycled by thermal or mechanical processes. Components with minor damages can be repaired. For other undamaged components, surface treatments or coating can be applied to increase the perceived quality. Therefore, it is important to evaluate the damage cases of these components, which can help to determine the correct repair or refurbish strategies in order to maintain the structure integrity. This will form our second case study, where HSI and computer vision will be used for surface damage detection and assessment.

In the manufacturing process, robotic systems become more and more widely adopted for industrial automation. For the industrial deployment of the HSI camera, an automatic HSI inspection system with human-robot interaction is proposed and trialled as our third case study. This system is integrated by a HSI camera and an industrial robotic manipulator, i.e., the KUKA KR06 R900, where the long arm of the robot can enable large-scale inspection of the components for satisfying real industrial needs.

The remaining paper is organised as follows: Section 2 describes the concept of the HSI based inspection system. Section 3 discusses the data acquisition and data processing of the system. Section 4 details the three case studies. Finally, some concluding remarks and future directions are summarised in Section 5.

2. HYPERSPECTRAL IMAGING INSPECTION SYSTEM

Fig. 3 depicts the concept of the hyperspectral imaging inspection system for the (re)manufacture industry, which is composed of three core modules, i.e. Imaging module, Machine-to-Machine (M2M) interaction Module, and Human-to-Machine (H2M) collaboration module. In the Imaging module, the optical sensors such as visible (VIS), near-infrared (NIR), and short-wave infrared (SWIR) hyperspectral cameras can provide a continuous spectral information over a certain spectral range, along with a good spatial resolution. M2M can actually provide a continuous information flow between the imaging devices and computer vision algorithms for data processing and analysis. Hereby, the sensors, onsite data acquisition machine, offsite data storage devices and cloud data processing units can connect and communicate to each other via the industrial IoT (mostly based on wireless network). In addition, H2M collaboration is also a very important module. Here, human workers, hyperspectral cameras and the specifically designed compliant robots can work together in carrying out complex and unstructured inspection tasks at the remanufacturing and repair production lines.

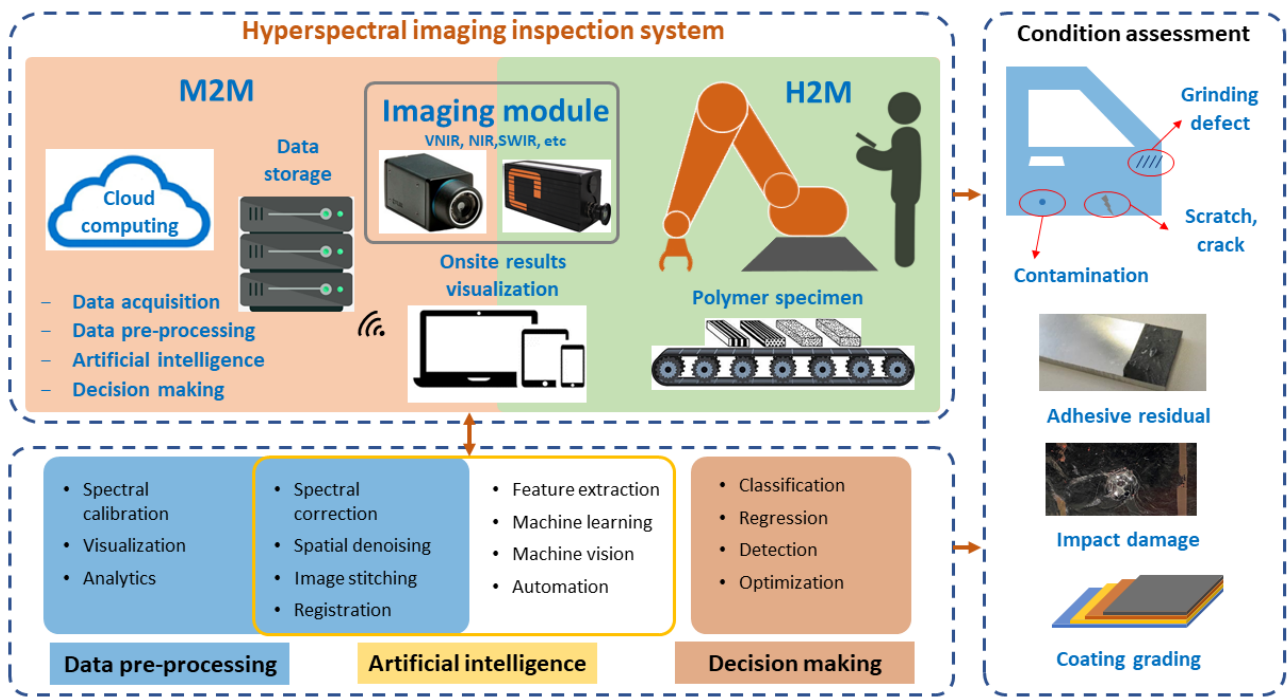

Fig. 3. Hyperspectral imaging inspection system for condition assessment in manufacturing and remanufacturing sectors, where M2M and H2M denote Machine-to-Machine and Human-to-Machine interactions, respectively.

Within the M2M techniques, data processing is considered as the enabling technology and covers three streams, i.e., data pre-processing, artificial intelligence (AI) and smart decision making. Data pre-processing is a vital step after data acquisition for data calibration and enhancement, which can produce high quality data for easy understanding and improved analysis via data visualization and analytics. According to the practical needs of different NDT tasks, data pre-processing sometimes needs to be incorporated into AI, where spectral correction and spatial denoising techniques are needed for improving the data quality, as well as image stitching and registration techniques for data fusion and full-scale representation.

The main AI techniques include but not limited to feature extraction, machine learning, computer vision and automation, etc. One of the biggest issues with analysing hyperspectral data is the large number of variables involved, leading to the curse of dimensionality or Hugh's phenomenon. A high number of variables necessitates a considerable number of samples thus the corresponding memory and processing capacity, and it may also lead to overfitting of the classifiers when training the model. To address this issue, feature extraction is needed for dimension reduction and more effectively characterisation of the data.

Machine learning algorithms have been widely applied in numerous industrial applications, which are usually built upon the sample data (also known as training data) before being utilised to make any estimates or predictions. The purpose of automation is to implement the monotonous tasks such as data acquisition by machines (i.e., KUKA robots), in order to reduce the human intervention and increase the efficiency and productivity.

Computer vision can be considered as an integration of feature extraction, machine learning and automation, which can provide imaging-based automatic inspection, big data process control and robotic implementation. The outcomes of these techniques are eventually used for decision making such as classification, regression, detection and optimization. The proposed hyperspectral imaging inspection system can be employed for generic condition monitoring and asset management for the repair and remanufacturing of CFRP materials. Relevant applications can be found in the detection of defects such as contamination, scratch, and grinding defect, recognition of adhesive residual, classification of impact damage types, and quality grading of coating.

3. TECHNIQUES AND METHODOLOGIES

In this section, the standard progress of data acquisition in the lab environment and data pre-processing techniques are introduced. Those techniques are not only used in this work but also applicable in many other HSI applications.

*3.1 Data acquisition*

In our three case studies, the hyperspectral data is acquired by two hyperspectral systems, where a visible (VIS) hyperspectral system is used for surface damage detection and a near-infrared (NIR) hyperspectral system for residual adhesive detection. Both systems can be incorporated into the robotic platform for H2M collaborations.

The two hyperspectral systems both operate in the push-broom mode, where the lens of the camera face downwards and scan only a single line at a time. Each scan produces a two-dimensional image representing the spatial line in one dimension and the full spectrum of each pixel in the other. Two 20W Tungsten halogen lights are used for illumination. For the first two case studies, the samples to be scanned are moved with a translational stage underneath the camera in a constant speed. With a working distance of 25cm, the speed of the translational stage is set to 16mm/s and the maximum length of the scanning path is 20cm to form a 3D hypercube. For the last case study, the translational stage is replaced by the KUKA robot arm, and the implementation detail will be discussed in Section 4.3.

The VIS imaging system contains a CCD camera, Hamamatsu ORCA-03G, and a spectrograph of Specim V8E. This system covers a spectral range of 400-950nm with a spectral resolution of 2nm. In addition, 4-fold spatial and spectral binning were applied in order to reduce the noise as well as to increase the camera's light sensitivity. This also results in an image with 336 pixels per line and each pixel contains 256 spectral responses.

The NIR imaging is done by the Innospec Red Eye 1.7 system covering a spectral range of 950-1700 nm with a spectral resolution of 10 nm. Without binning an image with 320 pixels per line and 256 active spectral bands is produced.

*3.2 Data pre-processing*

*3.1.1 Spectral calibration*

The illumination conditions may fluctuate in a hypercube or apparently among several datasets over the scan lines during the data acquisition procedure. To reduce the influence of difference in camera quantum and physical configuration of imaging systems, accurate calibrations for HSI system are necessary to guarantee the stability and acceptability of the extracted hyperspectral data. As a result, light calibration below is essential for converting the raw radiance spectrum $s$ to the reflectance spectra $r$ to eliminate such incoherence and maintain the effect of the light conditions consistent. Recording the camera's shot noise without any light exposure to the camera will give us a dark reference spectrum $d$, and recording an optimally reflective white surface (e.g., Spectralon with Lambertian scattering) will produce a white reference $w$. The light sensitivity to the current illumination can be estimated to normalise the signal by:

$$r = \frac{s - d}{w - d} \quad (1)$$

*3.1.2 Spectral correction*

As the extracted spectra depends not only on chemical absorption but also on physical light scattering, according to the objects' surface structure. Therefore, spectral correction is needed, after spectral calibration, to transfer the spectra to the Standard Normal Variate (SNV) [20] for minimising the errors caused by scattering. For any pixel with a reflectance spectrum $r_s$ in the HSI $D \in \Re^{I \times J \times B}$ at the location $(i, j)$ where $i \in 1:I, j \in 1:J$, the process of SNV is defined by:

$$r_s = D(i, j, b_{1:B}) \quad (2)$$

$$r_s(SNV) = \frac{r_s - \mu}{\sigma} \quad (3)$$

$$\mu = \frac{1}{I \times J} \sum_{s=1}^{I*J} r_s \quad (4)$$

$$\sigma = \sqrt{\frac{1}{I \times J} \sum_{s=1}^{I*J} (r_s - \mu)^2} \quad (5)$$

where $D(i, j, b_{1:B})$ represents a spectral vector with a length of $B$ at location $(i, j)$; $\mu$ and $\sigma$ denote the mean and standard deviation of all pixels in $r_s$.

*3.1.3 Spatial denoising*

Apart from the spectral degradation, the acquired data can also be degraded by instrumental noises such as thermal noise, quantisation noise and shot noise in the spatial domain. As a result, spatial denoising is needed to mitigate the noise. Herein, the bilateral filtering as a nonlinear approach that allows edge-preserved noise removal is employed. For a 3D hypercube, an improved joint bilateral filtering (JBF) [21] is used for spatially smoothing the data. Given a HSI data $D \in \Re^{I \times J \times B}$, I and J are the spatial size of $D$ and $B$ is the number of spectral bands. The JBF result of the input data $D$ at the location $(i,j)$ of band $b$ can be obtained by:

$$D_{JBF}(i, j, b) = \frac{1}{k(i,j)} \sum_{(p,q) \in w} \left( G_{\sigma_d}(i - p, j - q) \times G_{\sigma_r}(I_{PC1}(i,j) - I_{PC1}(p,q)) D(p,q,b) \right) \quad (6)$$

where $w$ is a local window centered at $(i, j)$ with a size of $(2\sigma_d + 1) \times (2\sigma_r + 1)$ pixels.

The normalisation factor $k$ is defined by

$$k(i,j) = \sum_{(p,q) \in w} \left( G_{\sigma_d}(i - p, j - q) \times G_{\sigma_r}(D_{PC1}(i,j) - D_{PC1}(p,q)) \right) \quad (7)$$

where $D_{PC1}$ is the first principal component derived from the principal component analysis (PCA), and $(p, q)$ denotes the spatial location of a pixel in the local window $w$.

The kernels for domain and range filtering are given by:

$$G_{\sigma_d}(i - p, j - q) = exp\left(-\frac{(i-p)^2 + (j-q)^2}{2\sigma_d^2}\right) \quad (8)$$

$$G_{\sigma_r}(D_{PC1}(i,j) - D_{PC1}(p,q)) = exp\left(-\frac{(D_{PC1}(i,j) - D_{PC1}(p,q))^2}{2\sigma_r^2}\right) \quad (9)$$

where $\sigma_d$ and $\sigma_r$ decide the neighbourhood window and the contributed weights of the pixels, respectively [22].

## 4. CASE STUDIES AND DISCUSSIONS

In this section, three case studies are presented to show the potential of proposed HSI based NDT of fibre materials, including specific AI techniques introduced in the context.

*4.1 Case study 1: Adhesive's residues detection*

*4.1.1 Experimental materials*

In the first study, three samples are used, which are single Aluminium (Al), adhesive bonded CFRP-CFRP and adhesive bonded CFRP-AI tensile shear samples (Table 1). CFRP composites are fabricated from automotive grade continuous fibre reinforced epoxy resin with the carbon fibres.

### 4.1.2 Experimental results

After data acquisition, two pre-processing steps are carried out, which include spectral calibration (Eq. (1)) and spectral correction (Eqs. (2-5)). To visualise the spectral characteristics of the surface conditions of the three samples, an image at the wavelength of 1267 nm is shown in Fig. 4, where the average spectral profiles over the whole bands of the selected regions of interest (ROI) are also platted for comparison in Fig. 4(a-c).

For the Al sample, three regions, i.e., normal region, grinding region and grinding defect region, are compared with different spectrums. For the normal region, the intensity is low as the surface is furbished and has a relatively lower reflectance. The lowest intensity on its normalised reflectance can be found between 1333nm and 1600nm. For the grinding region, it has high reflectivity, which leads to high reflectance and appears white on the colour image. Its normalised reflectance has the highest intensity between 1333nm and 1600nm, with a standard deviation lower than that of the other two regions in the range of 996-1333nm. Due to poor laser treatment, some defects are caused in the grinding region. The defect region has a lower reflectivity than the other grinding region, which looks grey on the colour image. Its normalised reflectance is in between of other two regions in the spectral range of 1333- 1600nm. Although its profile has the similar trend to the normal region within 996-1333nm, there are still some gaps between the two spectra, which shows their difference. Therefore, it can be concluded that NIR spectra is able to differ the furbished (normal) region, grinding region and grinding defect region.

In Fig. 4(b), four representative regions are highlighted in differently coloured boxes on the colour image. The blue and yellow boxes denote the normal region and the adhesive region on the Al side, respectively. The red and purple boxes are the normal region and the adhesive region on the CFRP side, respectively. In the colour image, it can hardly recognise the adhesive residual on the materials. However, apparent difference between the normal region and the adhesive region can be seen on the spectral profile. As seen in Fig. 4 (b), the wavelength 1147nm is an important turning point: the Al-normal region (blue line) has a lower reflectance before 1147nm but higher reflectance afterward than the Al-adhesive region (yellow line). On the contrary, opposite reflectance trends of the CFRP- normal region (red line) and the CFRP-adhesive region (purple line) are shown. In addition, the spectral profile of the Al-normal region (blue line) and the CFRP normal region (red line) can be well distinguished. The Al-adhesive region and the CFRP-adhesive region have similar spectral reflectance throughout the whole wavelength. These findings can validate the two capabilities of HSI, i.e., material identification and detection of adhesive residual.

Similarly, Fig. 4(c) presents four spectral profiles of two pairs of CFRP samples, which include two adhesive regions (yellow and purple boxes) and two normal regions (red and blue boxes). As can be seen, in each pair of samples, the spectral profiles show a high degree of similarity. However, the spectral profiles from the two pairs are much different, which indicate the inter-class dissimilarity between the normal and adhesive regions. Again, this finding also validates that HSI can potentially characterise different components/materials. Thus, it is tested for distinguishing the adhesive residual in this study, which can benefit the laser treatment during the disassembly stage and provide useful information for the grading of adhesive cleaning and other surface conditions.

### 4.2 Case study 2: Surface damage detection

#### 4.2.1 Experimental materials

In the second study, three kinds of CFRP specimens, manufactured by Resin Transfer Moulding (RTM) and Bulk Moulding Compound (BMC), were intentionally damaged by two types of impactors, while all the experiments are repeated three times. In the RTM process, fibre preform is laid in a two-part, matched, closed mould, before the thermoset is injected into the closed mould under low to moderate pressure

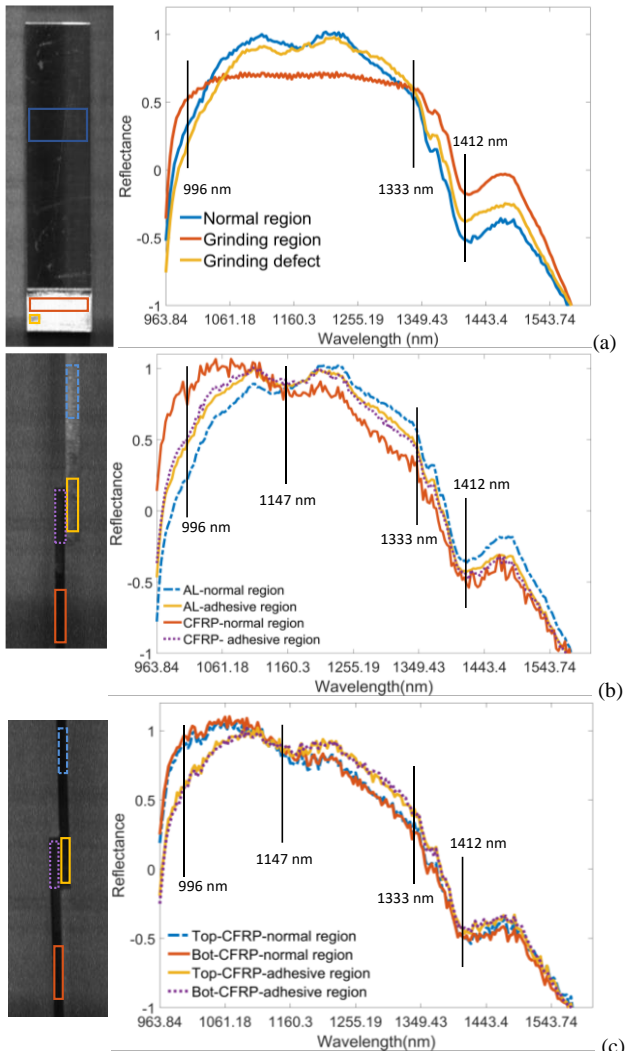

Fig. 4. Normalized spectra of different regions on the Al (a), CFRP+Al (b), and CFRP+CFRP (c). To the left are the colour images and to the right are the averaged spectrums from selected ROIs.

Table 1 Illustration of tensile shear samples and their dimension information.

| Material | Thickness (mm) |
| --- | --- |
| CFRP+CFRP | 4.48 |
| CFRP+Al | 5.6 |
| Al | 3.22 |

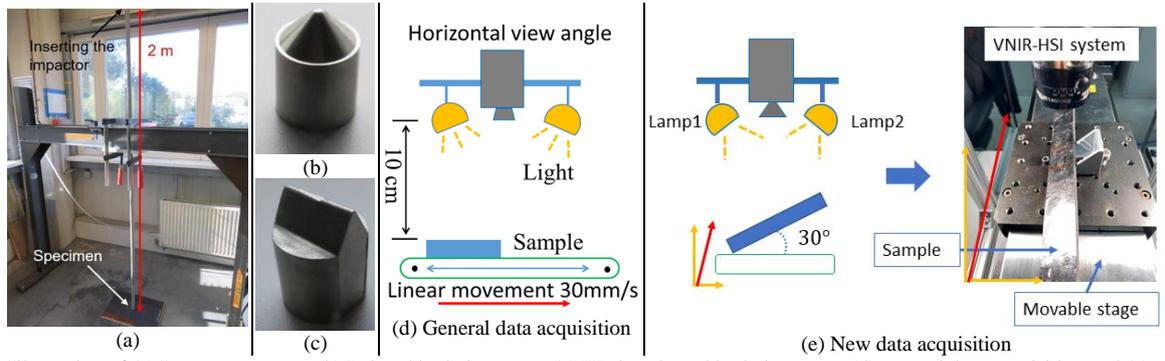

Fig. 5. Illustration of (a) Impact test setup, (b) Pointed body impactor, (c) Wedge-shaped body impactor, (d) general data acquisition and (e) new data acquisition in the second case study.

Table 2 Visual comparison between phone captured data (a), general (b) and new (c) data acquisition manners.

| | Point body impactor | | Wedge-shape body impactor | |
|---|---|---|---|---|
| (a) | | | | |
| (b) | | | | |
| (c) | | | | |

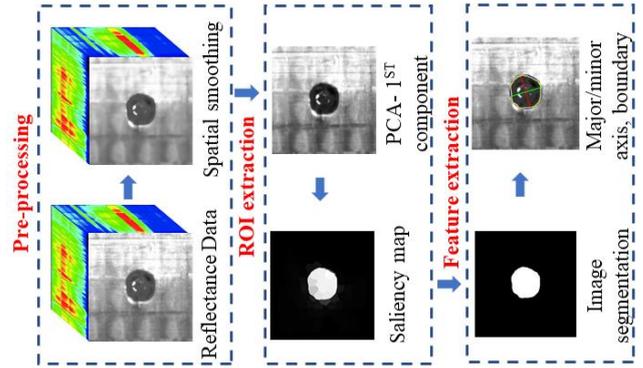

Fig. 6. Illustration of image processing framework.

[23]. For the BMC process, it is usually done by applying pressure and heat to the mould where a bulky mixture of fibres, resin paste and fillers is placed [24]. In our experiment, the first sample is manufactured via RTM using endless carbon fibre textile and epoxy matrix. The second sample is produced via BMC using new carbon long fibres, i.e., 45% fibre content with 25mm fibre length. The third sample is also manufactured by BMC using thermal recycled carbon long fibres i.e., 45% fibre content with 25mm fibre length. For a compact presentation, the three samples are named by RTM, BMC and rBMC respectively in the remaining paper. The experimental setup consists of a down pipe with a height of 2 meters. Each falling body has a weight of 607 g, which results in an impact energy of 11.9 joule. The impact test setup and results of the impact tests are shown in Fig. 5.

*4.2.2 Optimised data acquisition*

During the experiments, it is found that taking advantage of diffusion scattering and reflection of light can actually highlight the damaged regions even in the raw data. When the imaging camera vertically faces downwards the sample, the damaged region on the sample surface is hardly seen by the naked eyes. This is because the lights from two lamps form an even diffuse reflection on the surface of the sample when the image is acquired. Therefore, the camera may capture many unwanted textures on the surface, as shown in Table 2(b). By changing the placing angle between the sample and the movable stage, we can make the angle of incidence close to the angle of the reflection. This can bring in two benefits: 1) the camera can accurately record the light from lamp1 (Fig. 5(d)), and the surface can be captured in a good condition; 2) the diffusion scattering on the damaged region(s) can be enlarged so that the camera captures less light from those regions. On the other hand, lamp2 is also useful in this setting as it can further enlarge the diffusion scattering on the damaged region. Thus, the contrast between the damaged regions and the background can be enhanced in the captured data. Some damaged regions (Table 2(a)) that cannot be seen physically in conventional settings (Table 2(b)), now become clearly visible in the newly acquired data (Table 2(c)). The placing angle is dependent on the practical experimental settings, though we use 30° in our experiments. Afterwards, the acquired data will be translated back into the normal dimension by the cosine law. Some visualised results are shown in Table 2 to further illustrate how this process works.

*4.2.3 Damage detection framework*

Accurate distinguishing the types of damage is always of interest for determining the suitable repairing strategy for the damaged fibre composites. As such, an integrated HSI based damage detection system with machine learning and computer vision technologies is proposed. This is a multi-stage image processing framework (Fig. 6) that is composed of three key modules, i.e., pre-processing, saliency detection and feature output. It can detect the damaged regions and calculate the roundness ($Rd$) and rate of the major axis vs. the minor axis ($RMM$), see in Eqs. (10-11), as detailed below.

After data acquisition, two pre-processing steps are carried out, which include spectral calibration and spatial denoising, following the same procedure defined in Eq. (1) and Eqs. (6-9), respectively. Afterwards, ROI extraction is implemented from the first principal component of the hypercube through a saliency detection module [25] for generating a gray-scale saliency map of the damaged regions.

In the feature extraction step, a hard threshold is applied to convert the saliency map to a binary image, where the ROI will be highlighted as white pixels with the background in black. The boundary of ROI can be approximated by an ellipse with the associated parameters determined, including the roundness, length of the major axis and the minor axis that has the same normalised second central moments as the ROI.

| Material | (a) Visible data | (b) Ground truth | (c) Detected result | (d) Feature extraction | (e) Visible data | (f) Ground truth | (g) Detected result | (h) Feature extraction |
|---|---|---|---|---|---|---|---|---|
| RTM | | Precision | 98.07% | Recall | 99.49% | Precision | 87.46% | Recall | 21.71% |
| RTM | | Precision | 92.18% | Recall | 97.74% | Precision | 99.82% | Recall | 16.10% |
| RTM | | Precision | 98.66% | Recall | 91.43% | Precision | 63.47% | Recall | 29.87% |
| BMC | | Precision | 99.06% | Recall | 93.80% | Precision | 83.41% | Recall | 46.42% |
| BMC | | Precision | 98.77% | Recall | 94.54% | Precision | 96.11% | Recall | 28.06% |
| BMC | | Precision | 97.43% | Recall | 98.42% | Precision | 95.13% | Recall | 35.12% |
| RCY | | Precision | 99.66% | Recall | 82.07% | Precision | 97.67% | Recall | 65.27% |
| RCY | | Precision | 97.96% | Recall | 96.58% | Precision | 87.85% | Recall | 52.85% |
| RCY | | Precision | 93.25% | Recall | 97.86% | Precision | 77.11% | Recall | 75.80% |
| Overall | Precision | 97.23% | Recall | 94.66% | Precision | 87.56% | Recall | 41.24% |

Fig. 7. Illustrated results of damage detection: (a) damage (in red circle) caused by pointed body impactor, (b) ground truth of (a), (c) detection result of (a) and corresponding features (d); (e) damage (in red circle) caused by wedge-shaped body impactor, (f) ground truth of (e), (g) detection result of (e) and (h) corresponding parametric metrics.

$$Rd = \frac{a \cdot 4\pi}{p^2} \quad (10)$$

$$RMM = \frac{major\ axis}{minor\ axis} \quad (11)$$

where $a$ and $p$ denote the area and perimeter of the ROI, respectively.

### 4.2.4 Experimental results

To validate the usefulness of our damage detection system, we show examples of damage detection in Fig. 7, in which the colour images captured by a phone camera presented in columns (a) and (e) for visual comparison. The ground truth annotated by the domain experts are shown in columns (b) and (f). The damage detection results from the acquired HSI image are given in columns (c) and (g). The extracted features are illustrated on the HSI acquired data in columns (d) and (h), with the boundaries highlighted in yellow, the ellipse of the ROI shown in blue, and the major axis and the minor axis illustrated in red and green, respectively. In addition, the damage regions caused by the point body impactor and the wedge-shaped body impactor on the three materials are highlighted by red circles and red rectangles, respectively.

As shown in Fig. 7(a) and Fig. 7(e), the point body impactor causes a round-shape surface sag, and the wedge-shaped body impactor causes two bar-shape surface sag on the three materials. The bar-shape damage is hardly seen on the surface of RTM and BMC samples but is clear shown on the surface of RCY. This is because RTM and BMC are manufactured by original fibres, but RCY is manufactured by recycled BMC materials. As a result, the stiffness of RTM and BMC is higher than that of RCY. Another finding from Fig. 7 is that the damage region appears more clearly in the HSI than in the image from the phone camera, especially for the defects on the RTM and BMC materials. This has validated the efficacy of HSI than colour imaging in this context.

For quantitative evaluation, pixel-level precision and recall values are calculated when comparing the damage detection results with the manually defined ground truth as:

$$Precision = \frac{Tp}{Tp + Fp} \quad (12)$$

$$Recall = \frac{Tp}{Tp + Fn} \quad (13)$$

where $Tp$, $Fp$, and $Fn$ refer respectively to the number of correctly detected foreground pixels of the damage region, incorrectly detected foreground pixels (false alarms), and incorrectly detected background pixels (or missing pixels from the damage region) [25].

In the developed image processing framework, damages caused by the pointed body impactor can always be detected, with the precision and recall values over 90% (sometimes ~100%) in most cases. Nevertheless, it still has difficulty to detect damages caused by the wedge-shaped body as a whole on the surface of all materials. There are two main reasons: 1) the limited impact energy of 11.9 joule is not strong enough to make visible damages on the surface, 2) the complex nature of the surface texture has made the surface damage hardly visible. As a result, although the average detection precisions for BMC, RTM and RCY can reach 91.55%, 83.58% and 87.54%, the average recall values are only 36.53%, 22.56% and 64.64%, respectively. As seen in Fig. 7 (d), the damages on the RCY are often quite fully detectable because the recycled BMC material has a lower stiffness, making it more vulnerable to the impact and resulting in more visible surface damages. Therefore, its recall value is much higher than BMC and RTM.

Finally, the overall precision and recall measures for the two kinds of damages are obtained by the weighted average of all samples. For the pointed body damage and the wedge-shape body damage, the overall precision values on the three materials are 97.23% and 87.56%, respectively, with the overall recall values becoming 94.66% and 41.24%. Again, a high precision with a low recall value indicates that the HSI based integrated solution has a high reliability of positioning the impact damage. However, it fails to detect the whole region of the damage, especially for the wedge-shape body or complicated geometric damages.

After damage detection, the representative features of the damage regions i.e., roundness and the RMM can be extracted and plotted in Fig. 8 for visual comparison. With satisfied measures of the precision and recall, the RMM values for point body damage are consistently close to 1. The roundness varies from 0.65 to 1, and this is because the shape formed by the damaged region is not always round. For the wedge-shape body damages, due to a high precision and a low recall, both the RMM and roundness have large variance, where RMM ∈ [3.8, 8.8] and roundness ∈ [0.13, 0.35] . However, the features of these two damages are still very distinguishable, which again shows the capacity of HSI in NDT on the fibre material, and its potential for facilitating smart decision-making of requested repairing strategies.

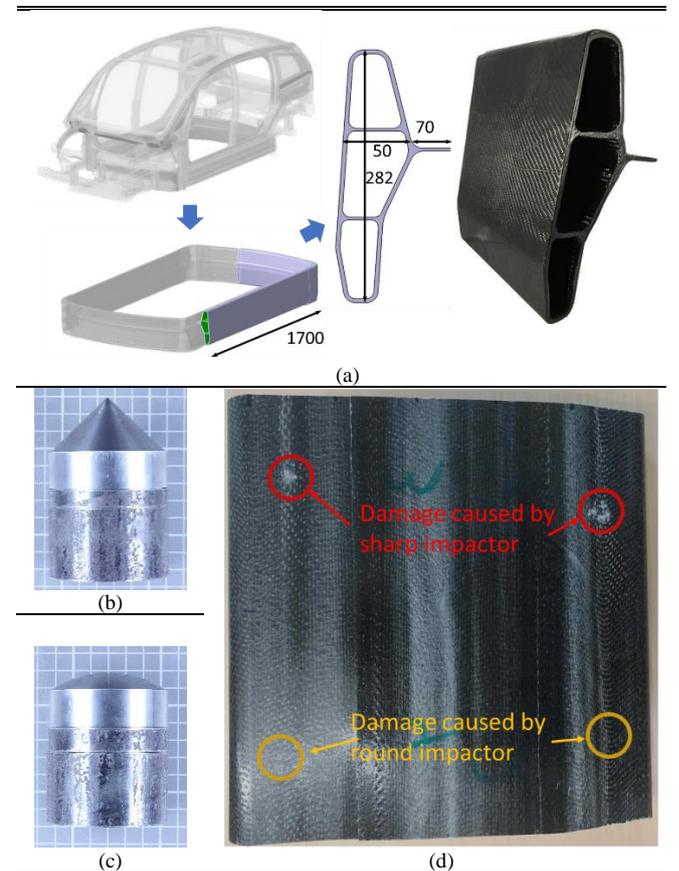

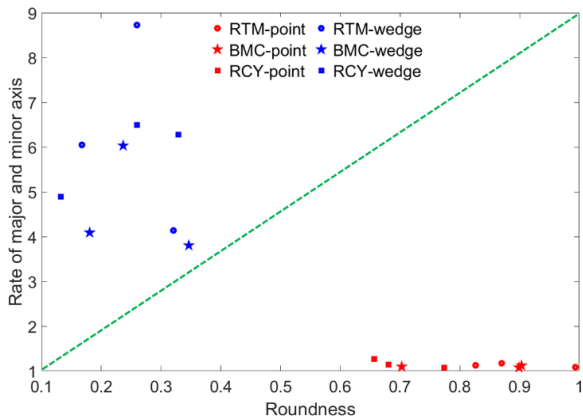

Fig. 8. Scatter plot of damage features.

Fig. 9. Illustration of the body structure (a), sharp impactor (b), round impactor (c), and sample surface with damages (d), all dimensions in mm.

| Parameters | Values |
|---|---|
| Weight | 55kg |
| Axis1 | ±170° |
| Axis2 | −190°/45° |
| Axis3 | −120°/156° |
| Axis4 | ±185° |
| Axis5 | ±120° |
| Axis6 | ±350° |

(a)
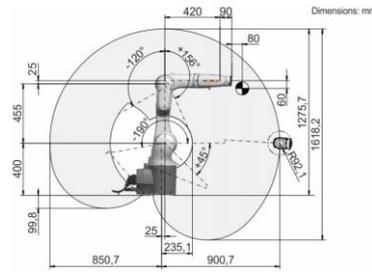
(b)
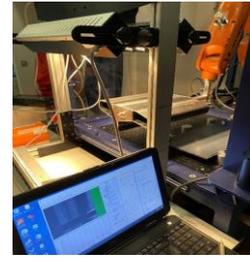
(c)

Fig. 10. Cobot system in the lab environment, (a) robot characteristics for KUKA KR06 R900, (b) working envelope with dimensions comprised during operation, (c) laboratorial cobot workplace.

*4.3 Case study 3: Cobot based automatic inspection*

*4.3.1 Experimental materials*

In the third study, a large automotive body structure sample that is made of UD-fibre (0°), biaxial fabric (0°/90°, −45°/45°) and a unsaturated polyester resin is damaged by a sharp impactor and a round impactor (Fig. 9). The experimental setup consists of a down tube with a height of 5.6 meters. The falling body has a mass of 2kg, which results in an impact energy of 109.8 joule. As seen in Fig. 9(d), the damage caused by the sharp impactor is noticeable, but the damage caused by round impactor is almost invisible.

*4.3.2 Experimental equipment*

In addition to the two HSI cameras, a robotic manipulator, the KR06 R900 [26] from KUKA is used in this case study, mainly for its universality in many industrial applications.. The robot is a 6-axis serial manipulator, with characteristics and working envelop shown in Fig. 10 (a) and (b).The real setting in the lab environment is shown in Fig. 10 (c), where the large automotive body structure is fixed to a holder mounted on the robot. The working distance between the camera and the sample surface is 25 cm. The movement of the robot is manipulated by an accompanying controller. During the camera scanning, the robot will drag the sample moving along Axis2 with a speed of $16 mm/s$. A laptop will record the hyperspectral data into the 'ENVI' format. The proposed system can also be implemented with any other serial manipulators such as Asea Brown Boveri (ABB) [27] and Stäubli [28].

*4.3.3 Image stitching*

Due to the limitation of the field of view, each scan can only cover a spatial dimension of 15 cm. As the width of the car body sample is 28 cm, dual scanning is needed to cover the whole surface. To represent the sample surface as a whole, a simple image stitching method is used in this study. Let the spatial dimension of scanned HSI data be $x * y$. The spatial dimension of the overall surface will be $(2x - \Delta x) * y$. In this study, x is decided by the spatial resolution of the HSI camera, which is 320 pixels. y is decided by the scanning path, or 620 pixels in total. $\Delta x$ is manually decided as 110 pixels in this study. To this end, the dimension of sample surface turns to be 530*620 pixels.

*4.3.4 Experimental results*

For visualisation purpose, image stitching is first applied to each HSI band, before conducting the PCA on the hypercube to extract the most representative features from the data. The 1st PCA component is shown in Fig. 11.

In Fig. 11(a), regions #1 and #2 highlight damages caused by sharp impactor, and regions #3 and #4 highlight damages caused by round impactor. In Fig. 11(b), regions #1-#4 are the corresponding healthy regions in Fig. 11(a). For the sharp impactor, it can break the fibre into the contacted region and deform the fibre in the surrounding regions. Therefore, it will make the damaged surface more sensitive to the light and

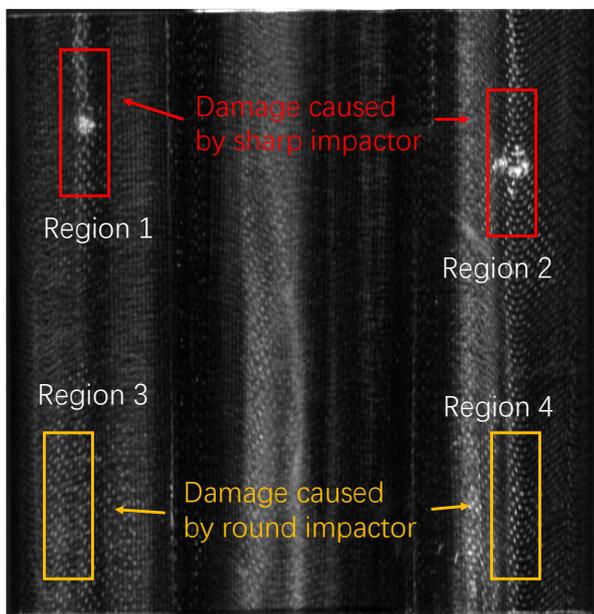
(a)
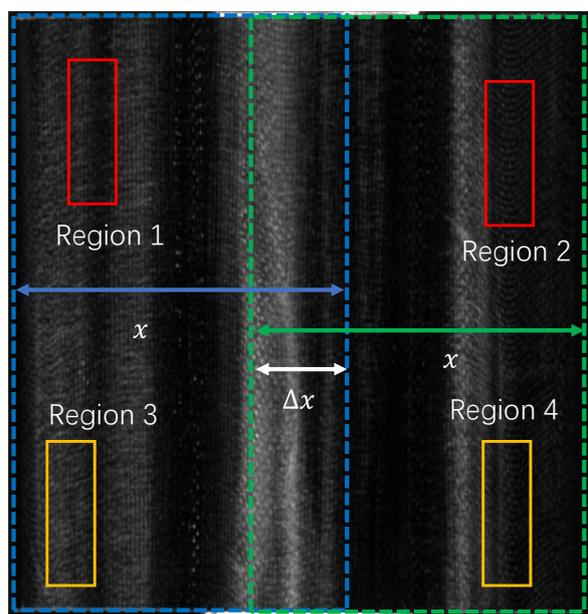
(b)

Fig. 11. Visualized 1st PCA component of sample front surface with damage (a) and without damage (b).

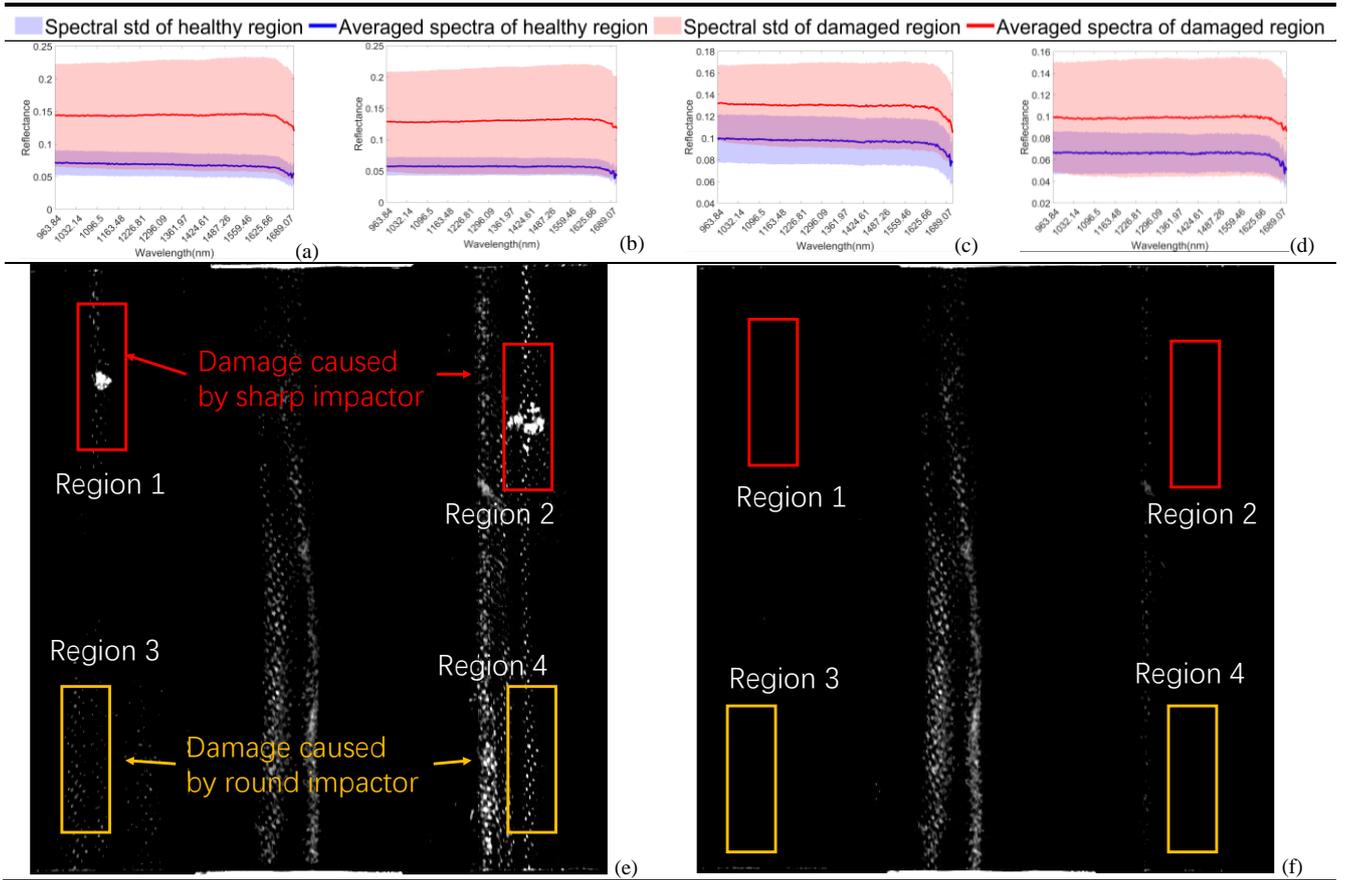

Fig. 12. Averaged spectral profile (a-d), and enhanced feature map of damaged regions (e) and healthy regions (f).

produce higher reflectance. As seen in Fig. 11, compared with healthy regions, impactor-contacted regions show a bright colour, with many white dots formed on deformed regions in the PCA domain. For round impactor, it won't break the fibre but will deform it in a larger area. As a result, there are only sparse white dots showing in the PCA domain.

To further highlight the difference between the damaged regions and healthy regions. We present averaged spectral profile for those regions in Fig. 12(a-d), where the standard deviation (std) is also plotted using shaded area. As the impactor will break and/or deform the fibre and change the reflectance of the neighbouring areas, both the reflectance and spectral standard deviation of the damaged regions are always higher than those of the healthy regions. By further suppressing the intensity of the background pixels, we can enhance the features in the PCA domain. As shown in Fig. 12, the pixel intensity in the healthy regions (Fig. 12(f)) appears darker than the damaged regions (Fig. 12 (e)) after the background suppression. On a different note, the damaged regions show bright colour on the feature map, which indicates obvious defects on the sample surface, again validates the efficacy of HSI in NDT of CFRP materials.

## 5. CONCLUSIONS AND FUTURE WORK

In this paper, novel insights are provided into HSI enabled non-destructive testing of CFRP materials, using three case studies on different kinds of CFRP materials. All the artificially introduced damages and failures follow the European standards. In the first case study, the adhesive residual shows a remarkable difference against the CFRP and aluminium materials on the spectral signature. In the second study, the combination of the HSI and computer vision techniques is able to well detect two types of surface damages on the CFRP materials. In the third case study, a human-robot collaboration platform is developed, where the HSI camera and robot arm are integrated together for automatic NDT. The major advantage of this platform is that it is a generic platform independent on any specified HSI sensors or robots. In theory, it can facilitate HSI-based NDT for any size of materials. The three case studies have fully shown the usefulness, effectiveness and great potential of HSI in NDT of CRFP and beyond.

In our future work, we will keep improving these three case studies for increased technology readiness level (TRL). First, how the amount of adhesive residual affects the spectral profile will be investigated, where a regression model can be built to quantitatively measure the concentration of adhesive residual through the HSI data. Second, the complex background is still the main obstruction for surface damage detection. As seen in Fig. 7 (h), the damage caused by the wedge-shaped body impactor cannot by fully detected. To solve this issue, some sophisticated background modelling methods [29] can be employed to eliminate the effect of the background for improved damage detection. Third, for the cobot system, we will design more tasks to test its flexibility and liability via in situ experiments. Apart from these studies, we will also explore new NDT tasks for improved adaptiveness and completeness to meet the ever-changing market requirements.

On a different note, there are still some challenges and limitations in this new NDT technique. First, it cannot fully detect the inner damage such as delamination and void. This limitation can be mitigated by integrating other NDT techniques such as ultrasonic testing and thermography. Second, the push-bloom style of imaging is time consuming, which will be a barrier to the practical application. This problem can be solved by using snapshot hyperspectral cameras such as Cubert [30] and Specim [31], etc.

ACKNOWLEDGEMENT

This work was supported in part by the EU H2020 grant 730323 (FibreEUse).